\documentclass[10pt,twocolumn,letterpaper]{article}

\usepackage{cvpr}
\usepackage{times}
\usepackage{epsfig}
\usepackage{graphicx}
\usepackage{amsmath}
\usepackage{amssymb}
\usepackage{bbm}
\usepackage{multirow}
\usepackage{comment}

\usepackage[dvipsnames]{xcolor}
\definecolor{pinkshade}{cmyk}{0, 0.7808, 0.4429, 0.1412}

\newcommand{\norm}[1]{\left\lVert#1\right\rVert}
\usepackage[pagebackref=true,breaklinks=true,letterpaper=true,colorlinks,bookmarks=false]{hyperref}

\usepackage[labelfont=bf, font=small]{caption} 

\cvprfinalcopy 

\begin{document}

\title{HandVoxNet: Deep Voxel-Based Network for 3D Hand Shape and Pose Estimation from a Single Depth Map\vspace{-0.3cm}} 

\author{
Jameel Malik$^{1,2,3}$ $\;\;\;\;\;$  
Ibrahim Abdelaziz$^{1,2}$  $\;\;\;\;\;$
Ahmed Elhayek$^{2,4}$ $\;\;\;\;\;$
Soshi Shimada$^{5}$\vspace{2pt}\\
$\;\;\,$Sk Aziz Ali$^{1,2}$  $\;\;\;\;\;\,$
$\;\;\,$Vladislav Golyanik$^{5}$  $\;\;\;\;\;$
Christian Theobalt$^{5}$ $\;\;\;\;$
Didier Stricker$^{1,2}$\vspace{7pt}\\
$^{1}$TU Kaiserslautern$\;\;$
$^{2}$DFKI Kaiserslautern$\;\;$
$^{3}$NUST Pakistan$\;\;$
$^{4}$UPM Saudi Arabia$\;\;$  
$^{5}$MPII Saarland$\;\;$ %
}

\maketitle

\begin{abstract}
3D hand shape and pose estimation from a single depth map is a new and challenging computer vision problem with many applications. 
The state-of-the-art methods directly regress 3D hand meshes from 2D depth images via 2D convolutional neural networks, which leads to artefacts in the estimations due to perspective distortions in the images. 

In contrast, we propose a novel architecture with 3D convolutions 
trained in a weakly-supervised manner. 
The input to our method is a 3D voxelized depth map, and we rely on two hand shape representations. 
The first one is the 3D voxelized grid of the shape which is accurate but does not preserve the mesh topology and the number of mesh vertices. 
The second representation is the 3D hand surface which is less accurate but does not suffer from the limitations of the first representation. 
We combine the advantages of these two representations by registering the hand surface to the voxelized hand shape. 
In the extensive experiments, the proposed approach improves over the state of the art by $47.8$\% on the \textit{SynHand5M} dataset. 
Moreover, our augmentation policy for voxelized depth maps further enhances the accuracy of 3D hand pose estimation on real data. 
Our method produces visually more reasonable and realistic hand shapes on  \textit{NYU} and \textit{BigHand2.2M} datasets compared to the existing  approaches. 
\end{abstract}

\section{Introduction}
\label{sec:Intro}
The problem of deep learning-based 3D hand pose estimation has been extensively studied in the past few years \cite{yuan2018depth}, and recent works achieve high accuracy on public benchmarks \cite{xiong2019a2j,moon2017v2v,rad2018feature}. 
Simultaneous estimation of 3D hand pose and shape from a single depth map is a newly emerging computer vision problem. 
It is more challenging than the pose estimation because
annotating real images for shape is laborious and cumbersome. 
Other salient challenges include varying hand shapes, occlusions, high number of degrees of freedom (DOF) and self-similarity. 
The dense 3D hand mesh is a richer representation which is more useful than the sparse 3D joints, and 
it finds many applications in computer vision and graphics \cite{taylor2016efficient,romero2017embodied,malik2019whsp}. 

\begin{figure}
    \centering
    \includegraphics[width=0.99\linewidth]{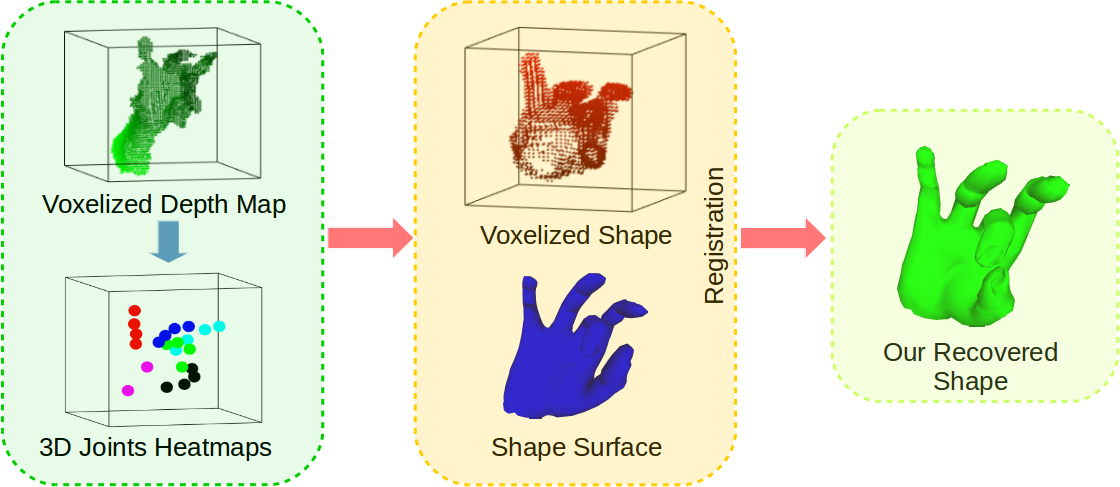}
    \vspace{3mm}
    \caption{
    \textbf{Hand shape and pose estimation with HandVoxNet.} 
    A 3D voxelized depth map and accurately regressed heatmaps of 3D joints (left block) are used to estimate two hand shape representations (middle block). 
    To combine the advantages of these representations, we accurately register the shape surface to the voxelized shape (right block). 
    Our architecture with 3D convolutions establishes a one-to-one mapping between voxelized depth map, voxelized hand shape and heatmaps of 3D joints. 
    } 
    \label{fig:HandVoxNet_Overview} 
\end{figure} 
With the recent progress in deep learning,
a few works \cite{zhang2019end,ge20193d,malik2019whsp,malik2019simple,malik2018deephps} have proposed algorithms for %
simultaneous hand pose and shape estimation. 
Malik \textit{et al.}~\cite{malik2018deephps} developed a 2D CNN-based approach that estimates shapes directly from 2D depth maps. 
The recovered shapes suffer from artifacts due to the limited representation capacity of their hand model 
\cite{ge20193d,malik2019whsp}. 
The same problem can occur even by embedding a realistic statistical hand model (\textit{i.e.,}~MANO) \cite{romero2017embodied} inside a deep network \cite{ge20193d,zhang2019end}. 
In contrast to these model-based approaches \cite{zhang2019end,malik2018deephps}, Ge \textit{et al.}~\cite{ge20193d} proposed a more accurate direct regression-based approach using a monocular RGB image. 
Recently, Malik \textit{et al.}~\cite{malik2019whsp} developed another direct regression-based approach from a single depth image. 
All of the approaches mentioned above treat and process depth maps with 2D CNNs, even though depth maps are intrinsically a 3D data. 
Training a 2D CNN to estimate 3D hand pose or shape given 2D representation of  a depth map is highly non-linear and results in perspective distortions in the  estimated outputs \cite{moon2017v2v}. 
V2V-PoseNet \cite{moon2017v2v} is the first work that uses 3D voxelized grid of depth map to estimate 3D joints heatmaps and, thus, avoids perspective distortions. 
However, extending this work for shape estimation by directly regressing 3D heatmaps of mesh vertices is not feasible in practice. 

\begin{figure*}[!ht]
\centering
\includegraphics[width=1\linewidth]{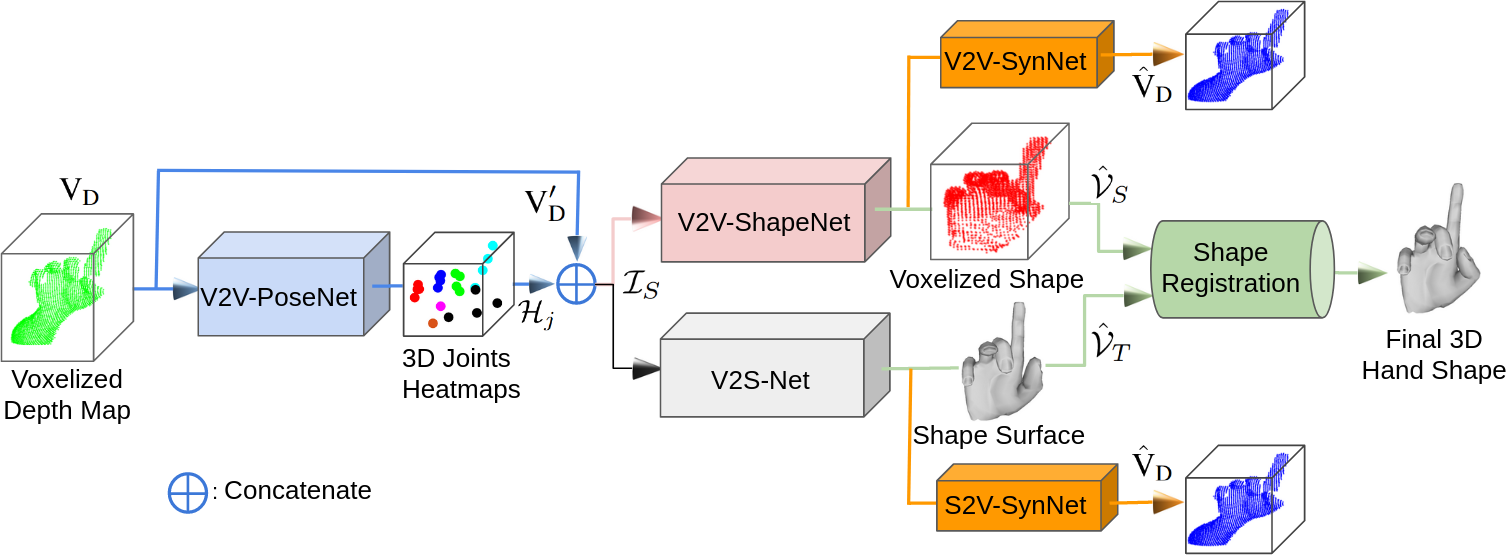}
\caption{
\textbf{Overview of our approach for 3D hand shape and pose recovery from a 3D voxelized depth map.} 
V2V-PoseNet estimates 3D joints heatmaps (\textit{i.e.,}~pose). Hand shape is obtained in two phases. First, V2V-ShapeNet and V2S-Net estimate the voxelized shape and shape surface, respectively. 
Thereby, V2V-SynNet and S2V-SynNet synthesize the voxelized depth acting as sources of weak-supervision. 
They are excluded during testing. 
In the second phase, shape registration accurately fits the shape surface to the voxelized shape.}
\label{fig:Pipeline}
\vspace{-2mm}
\end{figure*}

In this work, we propose the first 3D CNN architecture which simultaneously estimates 3D shape and 3D pose given a voxelized depth map (see 
Fig.~\ref{fig:HandVoxNet_Overview})
To this end, we introduce novel architectures based on 3D convolutions which estimate two different representations of hand shape (Secs.~\ref{sec:method_overview}--\ref{sec:NetTraining}). 
The first representation is the hand shape on a voxelized grid. 
It is estimated from a new \textit{voxel-to-voxel} network which establishes a one-to-one mapping between the voxelized depth map and the voxelized shape. 
However, the estimated voxelized shape does not preserve the hand mesh topology and the number of vertices. 
For this reason, we also estimate hand surface (the second representation) with our \textit{voxel-to-surface} network. 
Since this network does not establish a one-to-one mapping, the accuracy of the estimated hand surface is low but the hand topology is preserved. 
To combine the advantages of both representations, we propose  registration methods to fit the hand surface to voxelized shape. 
Since real hand shape annotations are not available, we employ two 3D CNN-based synthesizers which act as sources of weak supervision by generating voxelized depth maps from our shape representations (see Fig.~\ref{fig:Pipeline}). 
To increase the robustness and accuracy of the hand pose estimation, we perform 3D data augmentation on the voxelized depth maps (Sec.~\ref{ssec:DataAugmentation}). 

We conduct ablation studies and perform extensive evaluations of the proposed method on real and synthetic datasets. 
Our approach improves the accuracy of hand shape estimation by $47.8$\% on SynHand5M dataset \cite{malik2018deephps} and outperforms the state of the art. 
Our method produces visually more reasonable and plausible hand shapes of NYU and BigHand2.2M datasets compared to the state-of-the-art approaches  (Sec.~\ref{sec:experiments}). 
To summarise, our \textbf{contributions} are: 
\begin{enumerate}
  \item The first voxel-based hand shape and pose estimation approach with the following novel components: 
  
  (i) \textit{Voxel-to-voxel} 3D CNN-based network. 
  
  (ii) \textit{Voxel-to-surface} 3D CNN-based network. 
  
  (iii) 3D CNN-based voxelized depth map synthesizers.  

  (iv) Hand shape registration components. 
  \item A new 3D data augmentation policy on voxelized grids of depth maps. 
\end{enumerate}

\section{Related Work}\label{sec:related_work} 
We now discuss the existing methods for deep hand pose and shape  estimation. 
Moreover, we briefly report the most related works for depth-based hand pose  estimation. 
\paragraph{Deep Hand Pose and Shape Estimation.} 
Malik \textit{et al.}~\cite{malik2018deephps} proposed the first deep neural network for hand pose and shape estimation from a single depth image. 
To this end, 
they developed a model-based hand pose and shape layer which is embedded inside their deep network. 
Their approach suffers from artifacts due to the difficulty in optimizing complex hand shape parameters inside the network. 
Ge \textit{et al.}~\cite{ge20193d} developed a direct regression-based algorithm for hand pose and shape estimation from a single RGB image. 
They highlight that the representation capacity of the statistical deformable hand model (\textit{i.e.,}~MANO~\cite{romero2017embodied}) could be limited due to the small amount of training data and the linear bases utilized for the shape recovery. 
Zhang \textit{et al.}~\cite{zhang2019end} introduced a similar MANO model based approach using a monocular RGB image. 
Recently, Malik \textit{et al.}~\cite{malik2019whsp} proposed a structured weakly-supervised deep learning-based approach using a single depth image. 
All of the above-mentioned methods use 2D CNNs and treat the depth maps as 2D data. 
Consequently, the deep network is likely to produce perspective distortions in the shape and pose estimations \cite{moon2017v2v}. 
In contrast, we propose the first 3D convolutions based architecture which establishes a one-to-one mapping between the voxelized depth map and the voxelized hand shape. 
This one-to-one mapping allows to more accurately reconstruct the hand shapes.

\paragraph{Hand Pose Estimation from Depth.} 
In general, deep learning-based hand pose estimation methods can be classified into two categories. 
The first one encompasses the discriminative methods which directly estimate hand joint locations using CNNs  \cite{chen2019so,cai2019exploiting,xiong2019a2j,poier2019murauer,rad2018feature,moon2017v2v,guo2017region,ge2017robust,malik20183dairsig}.  
The second category is hybrid methods which explicitly incorporate hand structure inside deep networks \cite{malik2018structure,wan2018dense,ge2018point,oberweger2017deepprior++,malik2017simultaneous}. 
The disriminative methods achieve higher accuracy compared to the hybrid methods. 
The \textit{voxel-to-voxel} approach \cite{moon2017v2v} is powerful and highly effective because it uses 3D convolutions to learn a one-to-one mapping between the 3D voxelized depth map and 3D heatmaps of hand joints. 
Notably, the voxelized representation of depth maps is best suited for 3D data augmentation to improve the robustness and accuracy of the estimations.
A few methods perform data augmentation on depth maps~\cite{oberweger2017deepprior++,tompson2014real} or voxelized depth maps~\cite{moon2017v2v}.
In this work, we integrate the \textit{voxel-to-voxel} approach with our pipeline and, additionally, perform new 3D data augmentation on voxelized depth maps.     
Our 3D data augmentation policy helps to achieve a noticeable improvement in the 3D pose estimation accuracy on real datasets. 

\section{Method Overview}
\label{sec:method_overview}
Given a single input depth image, our goal is to estimate \textrm{N} 3D hand joint locations $\mathcal{J} \in \mathcal{R}^{3 \times \textrm{N}}$ (\textit{i.e.,}~3D pose) and $\textrm{K} = 1193$ 3D vertex locations $\mathcal{V} \in \mathcal{R}^{3 \times \textrm{K}}$ (\textit{i.e.,}~3D shape). Fig.~\ref{fig:Pipeline} shows an overview of the proposed approach. The input depth image is converted into a voxelized grid (\textit{i.e.,}~$\textrm{V}_{\textrm{D}}$) of size $88\times88\times88$, by using intrinsic camera parameters and a fixed cube size. 
For hand pose estimation, $\textrm{V}_\textrm{D}$ is provided as an input to the \textit{voxel-to-voxel} pose regression network (\textit{i.e.,}~V2V-PoseNet) that directly estimates 3D joint heatmaps $\{\mathcal{H}_j\}_{j=1}^\textrm{N}$. 
Each 3D joint heatmap is represented as $44\times44\times44$ voxelized grid.
We resize $\textrm{V}_\textrm{D}$ to $44\times44\times44$ voxel grid size (\textit{i.e.,}~${\textrm{V}}'_\textrm{D}$) and concatenate it with the estimated $\mathcal{H}_j$, to provide as an input to our shape estimation network. We call this concatenated input as $\mathcal{I}_S$.

The voxelized hand shape (\textit{i.e.,}~$64\times64\times64$ grid size) is directly regressed via 3D CNN-based \textit{voxel-to-voxel} shape regression network (\textit{i.e.,}~V2V-ShapeNet), by using $\mathcal{I}_S$ as an input. Notably, V2V-ShapeNet establishes a one-to-one mapping between the voxelized depth map and the voxelized shape. Therefore, it produces accurate voxelized shape representation but does not preserve the topology of hand mesh and the number of mesh vertices. 
To regress hand surface, $\mathcal{I}_S$ is fed to the 3D CNN-based \textit{voxel-to-surface} regression network (\textit{i.e.,}~V2S-Net). Since the mapping between
$\mathcal{I}_S$ and hand surface is not one-to-one, it is therefore less accurate. 
\textit{Voxel-to-voxel} and \textit{surface-to-voxel} synthesizers (\textit{i.e.,}~V2V-SynNet and S2V-SynNet) are connected after V2V-ShapeNet and V2S-Net, respectively. These synthesizers reconstruct ${\textrm{V}}'_\textrm{D}$ and  act as sources of weak supervision during training. 
They are excluded during testing. 
To combine the advantages of the two shape representations, we register the estimated hand surface to the estimated voxelized hand shape. We employ 3D CNN-based DispVoxNet \cite{ShimadaDispVoxNets2019} for synthetic data, and non-rigid gravitational approach (NRGA) \cite{Ali_NRGA_2018} for real data.

\section{The Proposed HandVoxNet Approach}
\label{sec:HandVoxNet}
In this section, we explain our proposed HandVoxNet approach by highlighting the function and effectiveness of each of its components. %
We develop an effective solution that produces reasonable hand shapes via 3D CNN-based deep networks. 
To this end, our approach fully exploits accurately estimated heatmaps of 3D joints as a strong pose prior, as well as voxelized depth maps. 
Given that collecting accurate real hand shape ground truth is hard and laborious, 
we develop a weakly-supervised network for real hand shape estimation by learning from accurately labeled synthetic data. 
Moreover,  our 3D data augmentation on voxelized depth maps 
allows to further improve the accuracy and robustness of 3D hand pose
estimation. 
\subsection{3D Hand Shape Estimation} 
\label{ssec:ShapeEstimation} 
As aforementioned, estimating 3D hand shape from a 2D depth map by using 2D CNN is a highly non-linear mapping. 
It compels the network to perform perspective-distortion-invariant estimation which causes difficulty in learning the shapes. To address this limitation, we develop a full voxel-based deep network that effectively utilizes the estimated 3D pose and voxelized depth map to produce reasonable 3D hand shapes. Our proposed approach for 3D shape estimation comprises of two main phases. In the first phase, we estimate the shape surface and the voxelized hand shape. In the second phase, we register the estimated shape surface to the estimated voxelized hand shape by employing a 3D CNN-based registration for synthetic data and NRGA-based fitting process for real data. 

\noindent\textbf{Voxelized Shape Estimation.}
Our idea is to estimate 3D hand shape in the voxelized form via 3D CNN-based network. It allows the network to estimate the shape in such a way that minimizes the chances for perspective distortion. Inspired by the approach proposed in the recent work \cite{malik2019whsp}, we consider sparse 3D joints as the latent representation of dense 3D shape. However, in this work, we combine 3D pose with the depth map which helps to represent the shape of hand more accurately. 
Furthermore, here we use more accurate and useful representations of 3D pose and 2D depth image which are 3D joints heatmaps and a voxelized depth map, respectively. The V2V-ShapeNet module is shown in Fig.~\ref{fig:Pipeline}. It can be considered as the 3D shape decoder: \begin{equation} \label{eq:0}
\hat{\mathcal{V}}_S \sim \textrm{Dec}( \mathcal{H}_j \oplus {\textrm{V}}'_\textrm{D}) = p(\mathcal{V}_S |\mathcal{I}_S)
\end{equation} 
where  $p(\mathcal{V}_S |\mathcal{I}_S)$ is the decoded distribution. The decoder learns to reconstruct the voxelized hand shape $\hat{\mathcal{V}}_S$ as close as possible to the ground truth voxelized hand shape $\mathcal{V}_S$.
The V2V-ShapeNet is a 3D CNN-based architecture that directly estimates the probability of each voxel in the voxelized shape indicating whether it is the background (\textit{i.e.,}~$0$) or the shape voxel (\textit{i.e.,}~$1$). 
The per-voxel binary cross entropy loss $\mathcal{L}_{\mathcal{V}_S}$ for voxelized shape reconstruction reads: 
\begin{equation} \label{eq:1}
\mathcal{L}_{\mathcal{V}_S} = - (\mathcal{V}_S\hspace{1mm}  \log(\hat{\mathcal{V}}_S) + (1-\mathcal{V}_S) \hspace{1mm}\log(1-\hat{\mathcal{V}}_S))
\end{equation}
where $\mathcal{V}_S$ and $\hat{\mathcal{V}}_S$ are
the ground truth and the estimated voxelized hand shapes, respectively. 
The architecture of V2V-ShapeNet is provided in the supplement. 

Since the annotations for real hand shapes are not available, weak supervision is therefore essential in order to effectively learn real hand shapes. For this reason, we propose a 3D CNN-based V2V-SynNet (see Fig.~\ref{fig:Pipeline}) which acts as a source of weak supervision during training. This module is removed during testing. V2V-SynNet synthesizes the voxelized depth map from the estimated voxelized shape representation. The per-voxel binary cross entropy loss $\mathcal{L}_{\textrm{V}_\textrm{D}}^\textrm{v}$ for voxelized depth map reconstruction is given by: 
\begin{equation} \label{eq:2}
\mathcal{L}_{\textrm{V}_\textrm{D}}^\textrm{v} = - (\textrm{V}_\textrm{D}\hspace{1mm}  \log(\hat{\textrm{V}}_\textrm{D}) + (1-\textrm{V}_\textrm{D}) \hspace{1mm}\log(1-\log(\hat{\textrm{V}}_\textrm{D}))
\end{equation}
where $\textrm{V}_\textrm{D}$ and $\hat{\textrm{V}}_\textrm{D}$ are %
the ground truth 
and the reconstructed voxelized depth maps, respectively.  
The architecture of V2V-SynNet is provided in the supplement.

\noindent\textbf{Shape Surface Estimation.} 
The hand poses of the shape surfaces and voxelized shapes need to be similar for an improved shape registration. 
To facilitate the registration, 
we employ V2S-Net deep network which directly regresses $\mathcal{V}$. Based on the similar concept of hand shape decoding (as mentioned before), $\mathcal{I}_S$ is provided as an input to this network while the decoded output is the reconstructed hand mesh (see Fig.~\ref{fig:Pipeline}). The hand shape surface reconstruction loss $\mathcal{L}_{\mathcal{V}_T}$ is given by the standard Euclidean loss as: 
\begin{equation} \label{eq:3}
\mathcal{L}_{\mathcal{V}_T} = \frac{1}{2}\norm{ \hat{\mathcal{V}}_T - \mathcal{V}_T }^2, 
\end{equation}
where $\mathcal{V}_T$ and $\hat{\mathcal{V}}_T$ are the respective ground truth and reconstructed hand shape surfaces. 
As explained before, in the case of missing real hand shape ground truth, the weak supervision on mesh vertices is provided by S2V-SynNet. 
In this case, the input to the S2V-SynNet is $\hat{\mathcal{V}}_T$ which is in  3D coordinates form. 
The loss function $\mathcal{L}_{\textrm{V}_\textrm{D}}^\textrm{s}$ for the S2V-SynNet is similar to Eq.~(\ref{eq:2}). 
Further details on S2V-SynNet and V2S-Net can be found in the supplement.  
\noindent\textbf{CNN-based Shape Registration.} 
Thanks to fully connected (FC) layers, V2S-Net is able to estimate hand shapes while preserving the order and number of points. Losing local spatial information is also known as a drawback of FC layers. In contrast to FC layers, a lot of works show fully convolutional networks (FCN) perform well in geometry regression tasks  \cite{shimada2019ismo,golyanik2018hdm,moon2017v2v,wu2016learning}. However, estimating the voxelized hand shape by 3D convolutional layer results in an inconsistent number of points and loses point order. 
Hence, the ideal architecture is a network which estimates the hand shape without losing local spatial information while preserving the topology of the hand shape. To achieve this, we register the estimated shape by V2S-Net to the probabilistic shape representation estimated by FCN (V2V-ShapeNet) using DispVoxNets pipeline \cite{ShimadaDispVoxNets2019}.  

The original DispVoxNets pipeline is comprised of 
two stages, \textit{i.e.,}~global displacement estimation and refinement stage. 
The refinement stage is used to remove roughness on the point set surface. In contrast to the original approach, we replace the refinement stage with Laplacian smoothing \cite{vollmer1999improved}. This is possible because we assume the mesh topology is already known, and it is preserved by our pipeline. 

In the DispVoxNet pipeline, the hand surface shape $\hat{\mathcal{V}}_T$ is first converted into a voxelized grid $\hat{\mathcal{V}}^{'}_T$ (\textit{i.e.,}~$64\times64\times64$ voxelized grid size). DispVoxNet estimates per-voxel displacements of the dimension $64^3\times3$ between the reference $\hat{\mathcal{V}}_S$ and voxelized hand surface $\hat{\mathcal{V}}^{'}_T$\footnote{a larger grid size results in higher accuracy of DispVoxNet, and we hence set it to the maximum which our hardware supports.}. 
The displacement loss $\mathcal{L}_{\textrm{Disp}}$ is given by: 
\begin{equation} 
\mathcal{L}_{Disp.}=\frac{1}{Q^{3}}\left\Vert \mathbf{d} - D_{vn}(\hat{\mathcal{V}}_S, \hat{\mathcal{V}}^{'}_T)\right\Vert^{2},
\end{equation}
where $Q$ and $\mathbf{d}$ are the voxel grid size and the ground truth displacement, respectively. Since it is difficult to obtain $\mathbf{d}$ between the voxelized shape $\hat{\mathcal{V}}_S$ and hand surface $\hat{\mathcal{V}}_T$, the displacements are first computed between $\mathcal{V}_T$ and $\hat{\mathcal{V}}_T$, and are discretized to obtain $\mathbf{d}$. 
For more details of ground truth voxelized grid computation, please refer to \cite{ShimadaDispVoxNets2019}. 
\vspace{0.25cm}

{\noindent\textbf{NRGA-Based Shape Registration.}} 
In our voxel-based 3D hand shape and pose estimation pipeline (Fig.~\ref{fig:Pipeline}), the DispVoxNet~\cite{ShimadaDispVoxNets2019} component requires shape annotations in its source-to-target displacement field learning phase. These annotations are available only for the synthetic dataset which leaves a domain gap on the performance of DispVoxNet when tested on real dataset. 
To bridge this gap, we apply NRGA~\cite{Ali_NRGA_2018} to improve $\hat{\mathcal{V}}_T$ by registering it with $\hat{\mathcal{V}}_S$. 
NRGA is selected for this deformable alignment task over other methods~\cite{papazov2011deformable,amberg2007optimal}, as it supports local topology preservation of input hand surface and is robust at noise handling. 
Although NRGA is a point cloud alignment method, it provides the option to relax the deformation magnitude in the neighbouring regions of the hand mesh vertices.
The original NRGA estimates a rigid transformation for every vertex $v \in \hat{\mathcal{V}}_T$ and diffuses the transformations in a subspace formed by a set of neighbourhood vertices of $v$. 
It builds a $k$-d tree on the template ($\hat{\mathcal{V}}_T$ in our case) and neighbourhood vertices are selected as the $k$-nearest neighbours (typically $0.1\% - 0.2\%$ of the total points in the template). 
We modify NRGA for the \textit{surface-to-voxel} operation, \textit{i.e.,} instead of $k$-nearest neighbours, we use connected vertices in a $4$-ring for $\hat{\mathcal{V}}_T$. See more details in our supplementary material. 
\subsection{Data Augmentation in 3D} 
\label{ssec:DataAugmentation} 
Our method for hand shape estimation relies on the accuracy of the estimated 3D pose. 
Therefore, the hand pose estimation method has to be accurate and robust.  
Training data augmentation helps to improve the performance of a deep network \cite{oberweger2017deepprior++}. 
Existing methods for hand pose estimation~\cite{oberweger2017deepprior++,tompson2014real} use data augmentation in 2D. 
This is mainly because these methods treat depth maps as 2D data. 
The representation of the depth map in voxelized form makes it convenient to perform data augmentation in all three dimensions. In this paper, we propose a new 3D data augmentation policy which improves the accuracy and robustness of hand pose estimation (see Sec.~\ref{ssec:EvalHandPose}). 

During V2V-PoseNet training, we apply simultaneous rotations in all three axes (\textrm{x},\textrm{y},\textrm{z}) to each 3D coordinate ($i,j,k$) of $\textrm{V}_\textrm{D}$ and $\mathcal{H}_j$ by using Euler transformations: 
\begin{equation} \label{eq:augmentation}
[\hat{i},\hat{j},\hat{k}]^\textrm{T} = 
[\textrm {Rot}_\textrm{x}(\theta_\textrm{x})]\times[\textrm {Rot}_\textrm{y}(\theta_\textrm{y})]\times
[\textrm {Rot}_\textrm{z}(\theta_\textrm{z})][i,j,k]^\textrm{T}, 
\end{equation}
where ($\hat{i},\hat{j},\hat{k}$) is the transformed voxel coordinate. 
$\textrm {Rot}_\textrm{x}(\theta_\textrm{x})$, $\textrm {Rot}_\textrm{y}(\theta_\textrm{y})$ and $\textrm {Rot}_\textrm{z}(\theta_\textrm{z})$ are $3\times3$ rotation matrices around $\textrm{x}$, $\textrm{y}$ and $\textrm{z}$ axes. 
The values for $\theta_\textrm{x}$, $\theta_\textrm{y}$ and $\theta_\textrm{z}$ are selected randomly in the ranges $[$$-40^{\circ}$, +$40^{\circ}$$]$, $[$$-40^{\circ}$, +$40^{\circ}$$]$ and $[$$-120^{\circ}$, +$120^{\circ}$$]$, respectively. 
In addition to rotations in 3D, following \cite{moon2017v2v}, we perform scaling and translation in the respective ranges $[+0.8, +1.2]$ and $[-8,+8]$. 

\vspace{7pt} 

\section{The Network Training} 

\label{sec:NetTraining}
$\textrm{V}_\textrm{D}$ is generated by projecting the raw depth image pixels into 3D space. Hand region points are then extracted by using a cube of size $300$ that is centered on hand palm center position. 
3D point coordinates of the hand region are discretized in the range $[$$1$, $88$$]$. 
Finally, to obtain $\textrm{V}_\textrm{D}$, the voxel value
is set to $1$ for the 3D point coordinate of hand region, and $0$ otherwise.  
Following \cite{moon2017v2v}, $\mathcal{H}_j$ are generated as 3D Gaussians. %
Similar to the generating of $\textrm{V}_\textrm{D}$, $\mathcal{V}_S$ is obtained by voxelizing the hand mesh. 
${\mathcal{V}}_T$  is created by normalizing the mesh vertices in the range $[$$-1$, $+1$$]$. 
We perform this normalization by
subtracting the vertices from the palm center and then dividing them by half of the cube size. 

We train V2V-PoseNet \cite{moon2017v2v} on NYU \cite{tompson2014real}, BigHand2.2M \cite{yuan2017bighand2} and SynHand5M \cite{malik2018deephps} datasets separately with the 3D data augmentation technique mentioned in Sec.~\ref{ssec:DataAugmentation}. 
For SynHand5M dataset, 
we train V2S-Net and V2V-ShapeNet (including the synthesizers S2V-SynNet and V2V-SynNet) separately using RMSProp as an optimization method with a batch size of $8$ and a learning rate LR = $2.5\times10^{-4}$. After training the pose and shape networks, we put these networks together in the pipeline (see Fig.~\ref{fig:Pipeline}) and fine-tune them in an end-to-end manner with synthetic, as well as combined real and synthetic data. The total loss $\mathcal{L}_\textrm{T}$ read as follows: 
\begin{equation}
    \mathcal{L}_\textrm{T} = \mathcal{L}_{\mathcal{H}} +
    \mathbbm{1}\mathcal{L}_{\mathcal{V}_S} +
    \mathbbm{1}\mathcal{L}_{\mathcal{V}_T} +
    \mathcal{L}_{\textrm{V}_\textrm{D}}^\textrm{v}
    +
    \mathcal{L}_{\textrm{V}_\textrm{D}}^\textrm{s}
\end{equation}
where $\mathcal{L}_{\mathcal{H}}$ is heatmaps loss \cite{moon2017v2v} and $\mathbbm{1}$ represents an indicator function layer. 
This layer forwards the estimations to the loss layer only for synthetic data using a flag value, which is $1$ for synthetic and $0$ for real data. 
It disables the gradients flow during the backward pass in the case of real data. 
For fine-tunings, we use RMSProp with a batch size of $6$ and a learning rate $2.5\times10^{-5}$. 
DispVoxNet is trained only on SynHand5M dataset due to the availability of the ground truth geometry. During the training, Adam optimizer \cite{adam} with a learning rate of $3.0 \times 10^{-4}$ was employed. The training continues until the convergence of $\mathcal{L}_{\textrm{Disp}}$ with batch size $12$. All models are trained until convergence on a desktop workstation equipped with Nvidia Titan X GPU. 

\begin{table}[t]
\begin{center}
\begin{tabular}{|l|c|}
\hline
Methods & 3D $\mathcal{V}$ Err. (\textit{mm}) \\
\hline\hline
V2S-Net (w/o $\mathcal{H}_j$) & 8.78 \\
V2S-Net (w/o ${\textrm{V}}'_\textrm{D}$) &  3.54 \\
V2S-Net (with $\mathcal{H}_j\oplus{\textrm{V}}'_\textrm{D}$) &  \textbf{3.36} \\
\hline
Methods & 3D $\mathcal{S}$ Err. \\
\hline
V2V-ShapeNet (w/o $\mathcal{H}_j$) &  0.007 \\
V2V-ShapeNet (w/o ${\textrm{V}}'_\textrm{D}$) & 0.016 \\
V2V-ShapeNet (with $\mathcal{H}_j\oplus{\textrm{V}}'_\textrm{D}$) &  \textbf{0.005} \\
\hline
\end{tabular}
\end{center}
\caption{\textbf{Ablation study on inputs (\textit{i.e.,}~$\mathcal{H}_j$ and ${\textrm{V}}'_\textrm{D}$) to V2S-Net and V2V-ShapeNet.} 
We observe that combining both inputs is useful for these two networks. 
}
\label{tab:AblationStudyConcatenation}
\end{table}

\begin{figure}[t]
\begin{center}
   \includegraphics[width=0.99\linewidth]{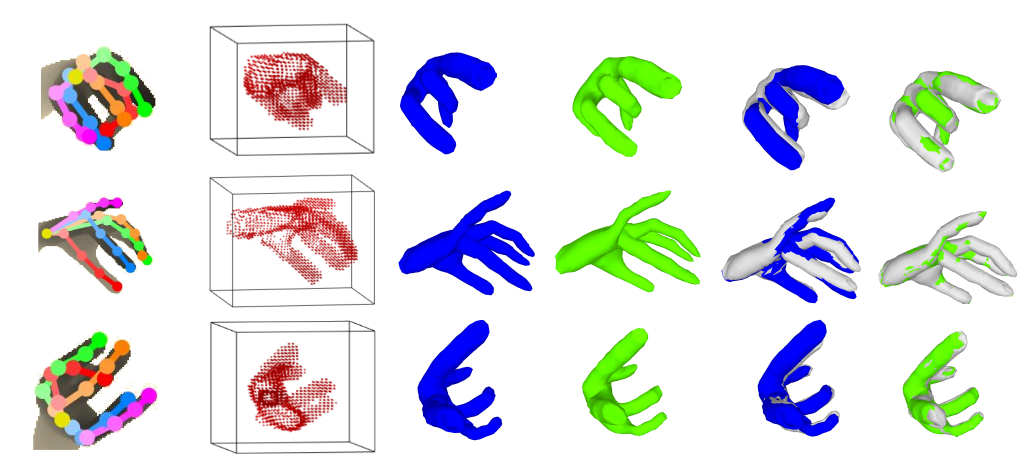}
\end{center}
   \vspace{-3mm}
   \caption{\textbf{Qualitative results on SynHand5M~\cite{malik2018deephps} dataset.} Estimated hand pose overlay ($1^\textrm{st}$ col), voxelized shape ($2^\textrm{nd}$ col), hand surface ($3^\textrm{rd}$ col), final shape ($4^\textrm{th}$ col), and the overlays of hand surface and final shapes with ground truth (gray color) are illustrated. 
   }
\label{fig:SynComparison_dvn}
\end{figure}

\begin{table}[t]
\begin{center}
\begin{tabular}{|l|c|}
\hline
Methods & 3D $\mathcal{V}$ Err. (\textit{mm}) \\
\hline\hline
DeepHPS \cite{malik2018deephps} & 11.8 \\
WHSP-Net \cite{malik2019whsp} & 5.12 \\
ours (w/o synthesizers) &  2.92 \\
ours (with synthesizers) &  \textbf{2.67}  \\
\hline
\end{tabular}
\end{center}
\caption{\textbf{Comparison with the state of the arts on SynHand5M \cite{malik2018deephps}.} Our full method, with V2V-SynNet and S2V-SynNet synthesizers, outperforms the WHSP-Net approach \cite{malik2019whsp} by $47.85$\%. 
}
\label{tab:ShapeResultsSynthetic}
\end{table}

\begin{figure*}[!ht]
\begin{center}
   \includegraphics[width=0.99\linewidth]{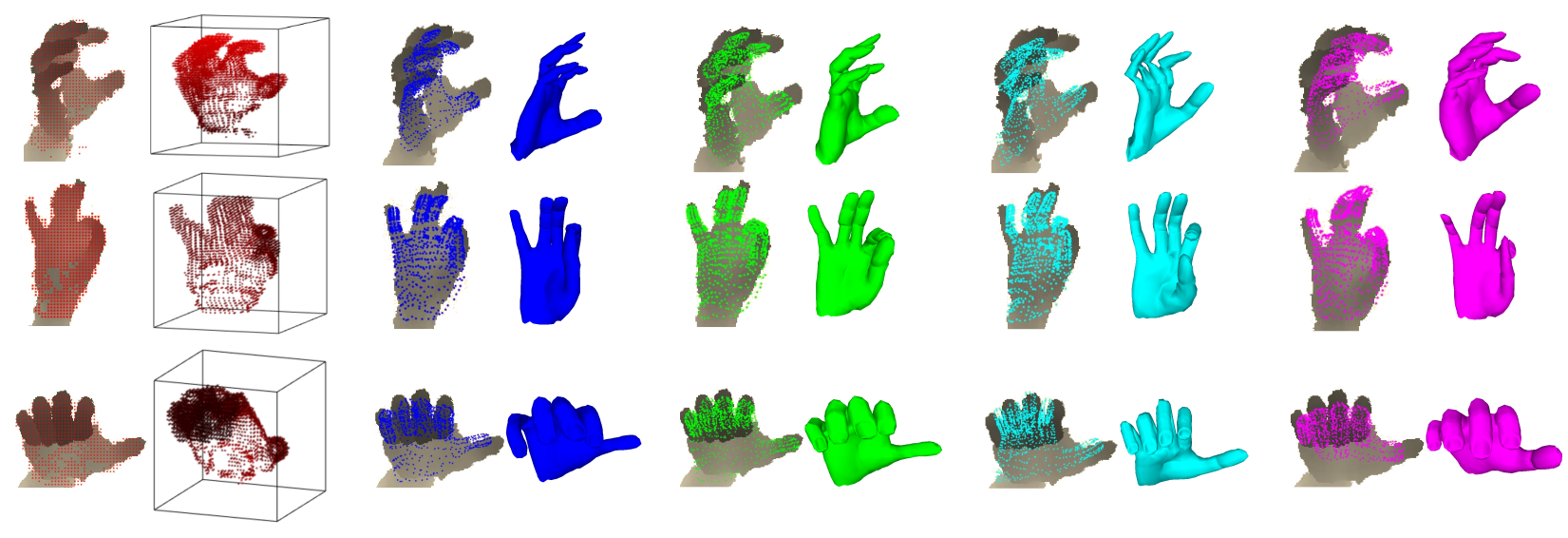}
   \vspace{-3mm}
   \rotatebox{0}{\hspace{2mm}(a) \small{Voxelized shape} \hspace{12mm} (b) \small{Shape surface} \hspace{9mm} (c) \small{Ours (final shape)} \hspace{5mm} (d) \small{DeepHPS \cite{malik2018deephps}} \hspace{5mm} (e) \small{WHSP-Net \cite{malik2019whsp}}}
\end{center}
   \caption{\textbf{Shape reconstruction of NYU \cite{tompson2014real} dataset}: (a), (b) and (c) show the 2D overlays and 3D visualizations of estimated voxelized hand shape, shape surface, and the final shape after registration, respectively. (d) and (e) show the corresponding results of hand shapes from DeepHPS \cite{malik2018deephps} and WHSP-Net \cite{malik2019whsp} methods. Our approach produces visually more accurate hand shapes than the existing approaches. 
   } 
\label{fig:NRGA_NYU}
\end{figure*}

\begin{figure}[t]
\begin{center}
   \includegraphics[width=0.99\linewidth]{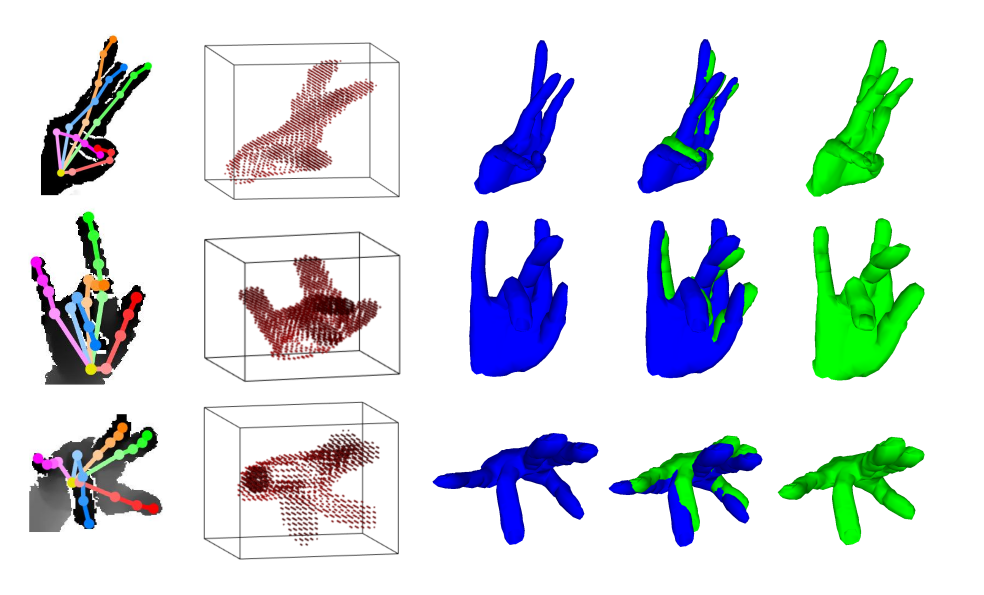}
   \rotatebox{0}{\hspace{1mm} (a) \hspace{12mm} (b) \hspace{13mm} (c) \hspace{10mm} (d) \hspace{8mm} (e) \hspace{7mm}}
\end{center}
   \vspace{-3mm}
   \caption{
   \textbf{Shape reconstruction of BigHand2.2M \cite{yuan2017bighand2} dataset:} (a) the 2D pose overlay; (b), (c) recovered voxelized shape and shape surface, respectively; (d) the overlays of shape surface and registered shape; (e) the final hand shape. 
   }
\label{fig:BIGHAND_POSE_COMPARISON}
\vspace{-5mm}
\end{figure}

\section{Experiments}
\label{sec:experiments}
We perform qualitative and quantitative evaluations of our complete pipeline
including ablation studies on the fully labeled SynHand5M \cite{malik2018deephps} dataset. 
We qualitatively evaluate real hand shape recovery on NYU  \cite{tompson2014real} and BigHand2.2M \cite{yuan2017bighand2} datasets. 
Furthermore, we study the impact of our 3D data augmentation on V2V-PoseNet  \cite{moon2017v2v}. 

\subsection{Datasets and Evaluation Metrics}
\label{ssec:metrics}
Although there are many depth-based hand pose datasets \cite{yuan2017bighand2,garcia2018first,supancic2015depth}, only a few of them (\textit{i.e.,}~BigHand2.2M \cite{yuan2017bighand2}, NYU \cite{tompson2014real}, SynHand5M \cite{malik2018deephps}) provide adequate training data and annotations which resemble the joint locations of a real hand. 
NYU real benchmark offers joint annotations for $72757$ and $8252$ RGBD images of the training ($\mathcal{T}_\textrm{N}$) and test sets, respectively.
Their hand model contains $42$ DOF
which makes it possible to combine this dataset with the recent benchmarks (\textit{e.g.,}~BigHand2.2M). 
BigHand2.2M is a million-scale real benchmark. For pose estimation, it provides accurate joint annotations for $956$\textrm{k} training ($\mathcal{T}_\textrm{B}$) depth images acquired from $10$ subjects. Their hand model contains $21$ joint locations which resembles real hand skeleton. The size of the BigHand2.2M's test set is $296$\textrm{k}. 
The annotation of hand palm center is not given in the BigHand2.2M dataset. Hence, we obtain the hand palm center position by taking the average of the metacarpal joints and the wrist joint positions. SynHand5M dataset contains fully annotated $5$ million depth images for both the 3D hand pose and shape. The sizes of its training ($\mathcal{T}_\textrm{S}$) and test sets are $4.5$\textrm{M} and $500$\textrm{k}, respectively. 
The joint annotations of BigHand2.2M are fully compatible with SynHand5M. 

We use three evaluation metrics: (i) the average 3D joint location error over all test frames (3D $\mathcal{J}$ Err.); (ii) mean vertex location error over all test frames (3D $\mathcal{V}$ Err.); and (iii) mean voxelized shape error (\textit{i.e.,}~per-voxel binary cross entropy) over all test data (3D $\mathcal{S}$  Err.).
\begin{table}[t]
\begin{center}
\begin{tabular}{|l|c|}
\hline
Components & runtime, \textit{sec} \\
\hline\hline
V2V-PoseNet &  0.011 \\
V2V-ShapeNet & 0.0015  \\
V2S-Net &   0.0038\\
DispVoxNet (GPU + CPU)$^*$ & 0.162 \\
NRGA (CPU)    & 59 - 70 \\
\hline
\end{tabular}
\end{center}
\caption{\textbf{Runtime:} (first four rows) forward-pass of deep networks on GPU. ``$^*$'' shows that Laplacian smoothing runs on CPU. 
}
\label{tab:Runtime}
\end{table}

\subsection{Evaluation of Hand Shape Estimation}
\label{ssec:EvalShapeEstim}
In this subsection, we evaluate our method on SynHand5M, NYU and BigHand2.2M benchmarks. 

{\noindent\textbf{Synthetic Hand Shape Reconstruction.}} 
We train our complete pipeline on the fully labeled SynHand5M dataset by following the training methodology explained in Sec.~\ref{sec:NetTraining}. 
We conduct two ablation studies to show the effectiveness of our design choice. 
First is the regression of $\mathcal{V}_T$ and $\mathcal{V}_S$ by 
using input ${\textrm{V}}'_\textrm{D}$ (\textit{i.e.,}~without  $\mathcal{H}_j$) and the synthesizers. 
Similar experiments are repeated by using $\mathcal{H}_j$ (\textit{i.e.,}~without ${\textrm{V}}'_\textrm{D}$) and $\mathcal{I}_S$ (\textit{i.e.,}~with $\mathcal{I}_S \oplus {\textrm{V}}'_\textrm{D}$) as separate inputs to V2V-ShapeNet and V2S-Net. 
The results are summarized in Table~\ref{tab:AblationStudyConcatenation}  and clearly show the benefit of concatenating voxelized depth map with 3D heatmaps. 
The second ablation study is to observe the impact of V2V-SynNet and S2V-SynNet, given $\mathcal{I}_S$ as an input to the complete shape estimation network. 
We train V2S-Net and V2V-ShapeNet with and without using their respective synthesizers (see Fig.~\ref{fig:Pipeline}). 
The quantitative results and comparisons with the state-of-the-art methods on SynHand5M test set are summarized in Table \ref{tab:ShapeResultsSynthetic}. 
Our method with synthesizers improves on ours without synthesizers, and achieves $47.8$\% improvement in the accuracy compared to the recent WHSP-Net \cite{malik2019whsp}. 
Several synthesized samples of voxelized depth maps are shown in the supplement. %
The qualitative results of shape representations and poses are shown in Fig.~\ref{fig:SynComparison_dvn}. 
DispVoxNet fits the estimated hand surface to the estimated voxelized hand shape, thereby improving the hand surface reconstruction accuracy by $20.5$\% (\textit{i.e.,}~ from $3.36mm$ to $2.67mm$). %
Notably,  the accuracy of our hand surface estimation is higher compared to WHSP-Net (\textit{cf.}~Tables  \ref{tab:AblationStudyConcatenation} and \ref{tab:ShapeResultsSynthetic}), which clearly shows the effectiveness of employing 3D CNN based network for mesh vertex regression. 

{\noindent\textbf{Real Hand Shape Reconstruction.}} 
To estimate plausible real hand shape representations, the synthesizers are essential (see Fig.~\ref{fig:Pipeline}). For NYU hand surface and voxelized shape recovery, we combine the training sets of NYU and SynHand5M (\textit{i.e.,}~$\mathbf{T}_{\textrm{NS}}=\mathcal{T}_\textrm{N}+\mathcal{T}_\textrm{S}$) by selecting closely matching $22$ common joint positions in both datasets. 
However, note that the common joint positions are still not exactly similar in both the datasets. 
V2S-Net and V2V-ShapeNet recover plausible hand shape representations while NRGA-based method performs a successful registration (as shown in Fig.~\ref{fig:NRGA_NYU}(a), (b) and (c)). 
It is observed that the voxelized shape is more accurately estimated than the hand surface. Thereby, the alignment further refines the hand surface. 
Using the similar training strategy, we combine BigHand2.2M and SynHand5M datasets and shuffle them (\textit{i.e.,}~$\mathbf{T}_{\textrm{BS}}=\mathcal{T}_\textrm{B}+\mathcal{T}_\textrm{S}$). Samples of the estimated hand shape representations for BigHand2.2M are shown in Fig.~\ref{fig:BIGHAND_POSE_COMPARISON}.  

We qualitatively compare our reconstructed hand shapes of NYU dataset with the state of the art. 
For better illustration of the shape reconstruction accuracy, we show the 2D overlay of hand mesh onto the corresponding depth image (as shown in Fig.~\ref{fig:NRGA_NYU}-(d) and (e)). 
Model-based DeepHPS \cite{malik2018deephps} suffers from artifacts, the  regression-based WHSP-Net approach \cite{malik2019whsp} produces perspective distortions and incorrect sizes of shapes. 
In contrast, HandVoxNet recovers visually more %
plausible hand shapes (Fig.~\ref{fig:NRGA_NYU}-(c)). 
Table~\ref{tab:Runtime} provides the runtimes of different components of our pipeline. 

\begin{figure}[t]
\begin{center}
   \includegraphics[width=1.0\linewidth]{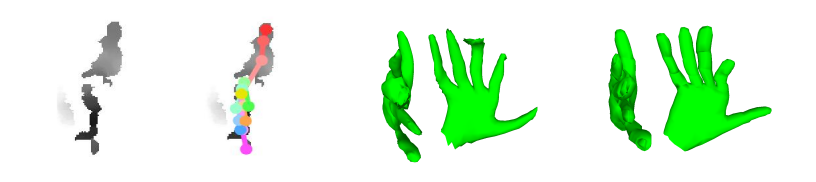}
   \rotatebox{0}{\hspace{0mm} (a) \small{Depth} \hspace{2mm}(b) \small{3D Pose} \hspace{2mm}(c) \small{DispVoxNet \cite{ShimadaDispVoxNets2019}} \hspace{3mm}(d) \small{NRGA \cite{Ali_NRGA_2018}}\hspace{5mm}}
\end{center}
   \vspace{-3mm}
   \caption{%
   \textbf{Failure case}: our method is unable to produce plausible shapes in cases of severe occlusion and missing depth information.} 
   \label{fig:failure_dvn}
\end{figure}

\begin{table}[t]
\begin{center}
\begin{tabular}{|l|c|}
\hline
Methods & 3D $\mathcal{J}$ Err. (\textit{mm}) \\
\hline\hline
DeepHPS \cite{malik2018deephps} & 6.30 \\
WHSP-Net \cite{malik2019whsp} & 4.32 \\
V2V-PoseNet \cite{moon2017v2v} &  3.81 \\
our HandVoxNet (full method) &  \textbf{3.75}\\
\hline
\end{tabular}
\end{center}
\caption{\textbf{3D hand pose estimation results on SynHand5M \cite{malik2018deephps} dataset.} We compare the accuracy of our full method (\textit{i.e.,}~HandVoxNet) with state-of-the-art methods.}
\label{tab:PoseResultsSynthetic}
\end{table}

{\noindent\textbf{Failure Cases.}} 
Our approach fails to estimate plausible hand shapes in cases of severe occlusion of hand parts and missing information in the depth map (see Fig.~\ref{fig:failure_dvn}). 

\begin{table}[t]
\footnotesize
\centering
\begin{tabular}{|c|c|c|}
\hline
Dataset & Method & 3D $\mathcal{J}$ Err. (\textit{mm}) \\ \hline \hline
\multirow{2}{*}{NYU} & V2V-PoseNet~\cite{moon2017v2v} & 9.22 \\
 & V2V-PoseNet (our 3D augm.) & \textbf{8.72} \\
\hline
\multirow{2}{*}{BigHand2.2M} & V2V-PoseNet~\cite{moon2017v2v} & 9.95 \\
 & V2V-PoseNet (our 3D augm.) & \textbf{9.27} \\
\hline
\end{tabular}
\vspace{3mm}
\caption{\textbf{3D hand pose estimation results on NYU \cite{tompson2014real} and BigHand2.2M \cite{yuan2017bighand2} datasets} using our 3D data augmentation.} 
\label{tab:PoseResultsNYU}
\vspace{-2mm}
\end{table} 

\begin{figure}[t]
\begin{center}
   \rotatebox{90}{\hspace{5mm} \small{Ours} \hspace{8mm} \small{WHSP-Net} \hspace{5mm} \small{DeepHPS}}
   \includegraphics[width=0.95\linewidth]{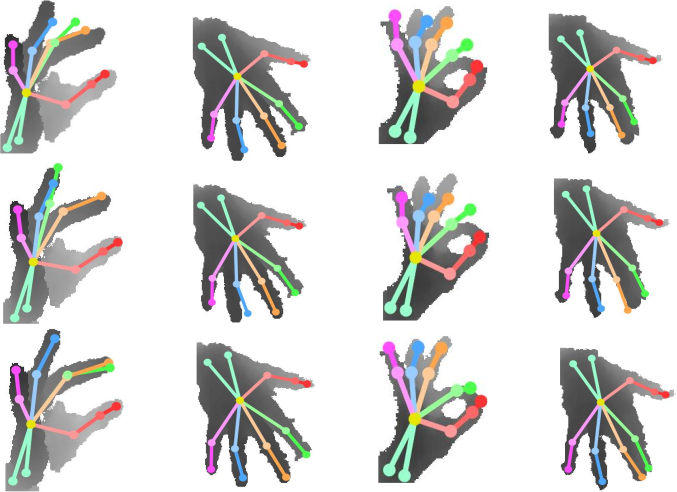}
\end{center}
   \vspace{-3mm}
   \caption{\textbf{Samples of NYU \cite{tompson2014real} depth images} with 2D overlay of the estimated 3D hand pose. Our method produces more accurate results compared to WHSP-Net \cite{malik2019whsp} and DeepHPS \cite{malik2018deephps} methods. 
   }
\label{fig:NYU_POSE_COMPARISON}
\vspace{-5mm}
\end{figure}

\subsection{Evaluation of Hand Pose Estimation} 
\label{ssec:EvalHandPose} 
In our approach, the accuracy of the estimated hand shape is dependent on the accuracy of estimated 3D pose (see Sec.~\ref{sec:HandVoxNet}). 
Therefore, the hand pose estimation needs to be robust and accurate. 
Therefore, we perform a new 3D data augmentation on voxelized depth maps which further improves the accuracy of 3D hand pose estimation on real datasets. 
Notably, our focus is to develop an effective approach for simultaneous hand pose and shape estimation. However, for completeness, we show our results and comparisons of hand pose estimation with SynHand5M, NYU and BigHand2.2M datasets. 

{\noindent\textbf{SynHand5M dataset:}} 
We do not perform training data augmentation on SynHand5M because this dataset originally contains large viewpoint variations \cite{malik2018deephps}.
We train our full method and V2V-PoseNet \cite{moon2017v2v} on SynHand5M dataset.
The quantitative results on the test set are presented in Table \ref{tab:PoseResultsSynthetic}. 
We observe that the backpropagation from the shape regression pipeline is effective and improves the accuracy of the estimated 3D pose. 
We achieve $13.19$\% improvement in the accuracy compared to WHSP-Net  approach \cite{malik2019whsp}. 

\begin{figure}[t]
\begin{center}
  \includegraphics[width=0.99\linewidth]{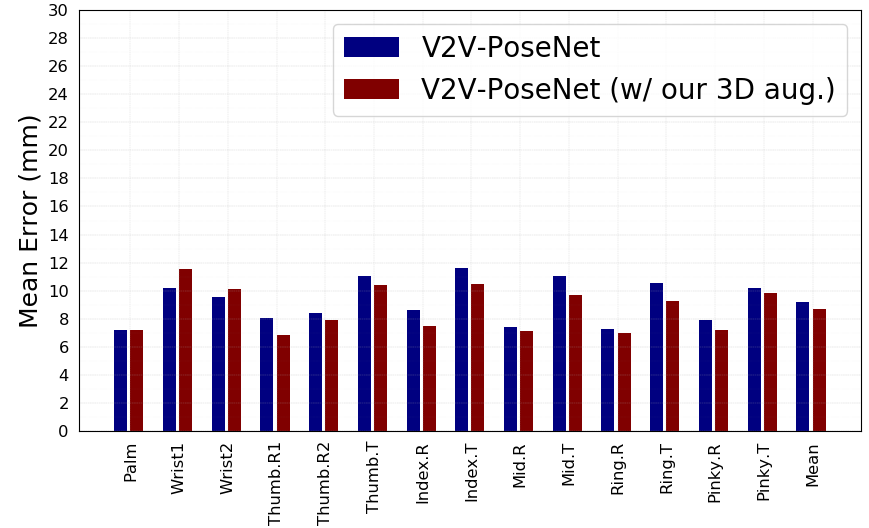}
\end{center}
  \vspace{-3mm}
  \caption{
  \textbf{We study the impact of our 3D data augmentation} on the pose estimation accuracy of V2V-PoseNet \cite{moon2017v2v} on NYU \cite{tompson2014real} dataset. The graph shows mean errors for  individual hand joints. 
  } 
\label{fig:NYU_POSE_GRPAH}
\vspace{-5mm}
\end{figure}

{\noindent\textbf{NYU and BigHand2.2M datasets:}} 
V2V-PoseNet \cite{moon2017v2v} is a powerful pose estimation method that exploits the 3D data representations of hand pose and depth map. 
Thanks to our 3D data augmentation strategy (see  Sec.~\ref{ssec:DataAugmentation}), we improve the accuracy by $5.42$\% and $6.83$\% compared to the original V2V-PoseNet models on NYU and BigHand2.2M datasets, respectively (see Table \ref{tab:PoseResultsNYU}). 
Fig.~\ref{fig:NYU_POSE_GRPAH} shows the average errors on individual hand joints. 
We observe a noticeable improvement in the accuracy of the finger tips. 
The qualitative results and comparisons with the state-of-the-art methods for hand pose estimation are shown in Fig.~\ref{fig:NYU_POSE_COMPARISON}. 

\section{Conclusion and Future Work}
We develop the first voxel-based pipeline for 3D hand shape and pose recovery  from a single depth map, which establishes an effective inter-link between hand pose and shape estimations using 3D convolutions. 
This inter-link boosts the accuracy of both estimates, which is demonstrated experimentally. 
We employ 3D voxelized depth map and accurately estimated 3D heatmaps of joints  as inputs to reconstruct two hand shape representations, \textit{i.e.,}~3D voxelized shape and 3D shape surface. 
To combine the advantages of both shape representations, we employ  registration methods, \textit{i.e.,} DispVoxNet and NRGA, which accurately fit the shape surface to the voxelized shape. 

The experimental evaluation further shows that our 3D data augmentation policy on voxelized grids enhances the accuracy of 3D hand pose estimation on real data. 
HandVoxNet produces visually more accurate hand shapes of real images compared to the previous methods. 
All these results indicate that the one-to-one mapping between voxelized depth map, voxelized shape and 3D heatmaps of joints is essential for an accurate hand shape and pose recovery. 

In future work, generating a realistic synthetic dataset can further enhance the hand shape reconstruction from real images. 
The runtimes of the used registration methods can be improved by the parallelization on GPUs. 

{\textbf{Acknowledgement:}} This work was funded by the German Federal Ministry of Education and Research as part of the project VIDETE (grant number 01IW18002) and the ERC Consolidator Grant 770784. 

{\small
\bibliographystyle{ieee_fullname}
\bibliography{egbib}
}

\end{document}